\documentclass[10pt,letterpaper]{article}

\usepackage{iccv}
\usepackage{times}
\usepackage{epsfig}
\usepackage{graphicx}
\usepackage{amsmath}
\usepackage{amssymb}
\usepackage{caption}
\usepackage{subcaption}
\usepackage{booktabs}
\usepackage{array}

\usepackage[breaklinks=true,bookmarks=false]{hyperref}

\iccvfinalcopy 

\def\iccvPaperID{****} 

\ificcvfinal\fi

\begin{document}

\title{Depth-NeuS: Neural Implicit Surfaces Learning for Multi-view Reconstruction Based on Depth Information Optimization}
\author{
Hanqi Jiang\qquad Cheng Zeng\qquad Runnan Chen \\
Beijing Jiaotong University \\
Beijing, China \\
{\tt\small \{20722010,20726043,20711083\}@bjtu.edu.cn}
\and
Shuai Liang$^{\dagger}$\qquad Yinhe Han\qquad Yichao Gao\qquad Conglin Wang \\
Institute of Computing Technology, Chinese Academy of Sciences\\
Beijing, China\\
{\tt\small \{liangshuai,yinhes,gaoyichao,wangconglin\}@ict.ac.cn}
}

\maketitle
\ificcvfinal\thispagestyle{empty}\fi

\begin{abstract}
   Recently, methods for neural surface representation and rendering, for example NeuS, have shown that learning neural implicit surfaces through volume rendering is becoming increasingly popular and making good progress. However, these methods still face some challenges. Existing methods lack a direct representation of depth information, which makes object reconstruction unrestricted by geometric features, resulting in poor reconstruction of objects with texture and color features. This is because existing methods only use surface normals to represent implicit surfaces without using depth information. Therefore, these methods cannot model the detailed surface features of objects well. To address this problem, we propose a neural implicit surface learning method called Depth-NeuS based on depth information optimization for multi-view reconstruction. In this paper, we introduce depth loss to explicitly constrain SDF regression and introduce geometric consistency loss to optimize for low-texture areas. Specific experiments show that Depth-NeuS outperforms existing technologies in multiple scenarios and achieves high-quality surface reconstruction in multiple scenarios.
\end{abstract}


\section{Introduction}
In computer vision field, the reconstruction of 3D scenes from pixel-wise images becomes a significant research direction. And this topic brings many practical insights into industries: manufacturing model reformulation, urban area planning, robotics, etc. The groundworks regarding this topic propose fixed shape representation techniques to better fit the geometry appearance. However, these methods cannot provide more precise reconstruction outputs, due to their limited ability to fit continuous shape in 3D world.
With the emergence of implicit neural surface representation, the best-practice of parameterization has been established. The state-of-art approaches of parameterization works include the two adaptations: volumetric and surface. Recent works~\cite{IM-Net,neus,DVR,neural} elaborate on the inherent differentiable nature of volumetric representation. Although regular voxel grids parametrization can achieve highly extensive performance on learning 3D scenes from inverse images, they can consume much computational capabilities and memories, and their accuracy are not desirable. Meanwhile, the surface implicit representation builds up a more intuitive network structure. Many works~\cite{2020deep,gropp2020implicit,deepsdf,metasdf} use pre-trained additional features, such as SDF function values to assist the learning process. The latter method now raises more recognition, since it allows the model to learn the intrinsic features of the scene, and the training outcomes often can easily attain a much higher score in terms of accuracy.

\begin{figure*}[htbp] 
\centering
\begin{minipage}[t]{0.33\linewidth} 
\centering
\includegraphics[width=0.95\linewidth]{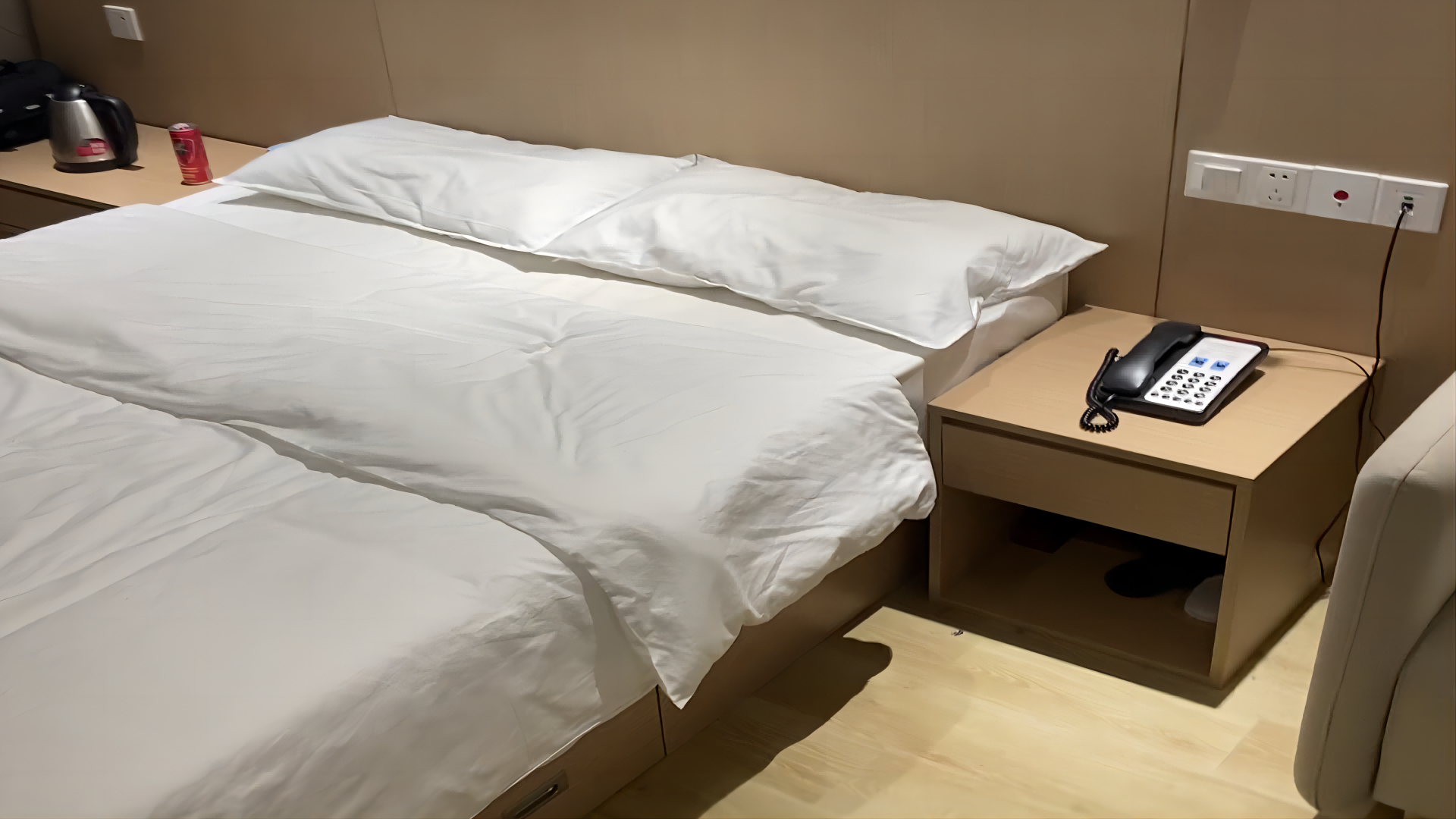} 
\subcaption{\label{fig1}{\small Reference Image}} 
\end{minipage}%
\begin{minipage}[t]{0.33\linewidth}
\centering
\includegraphics[width=0.95\linewidth]{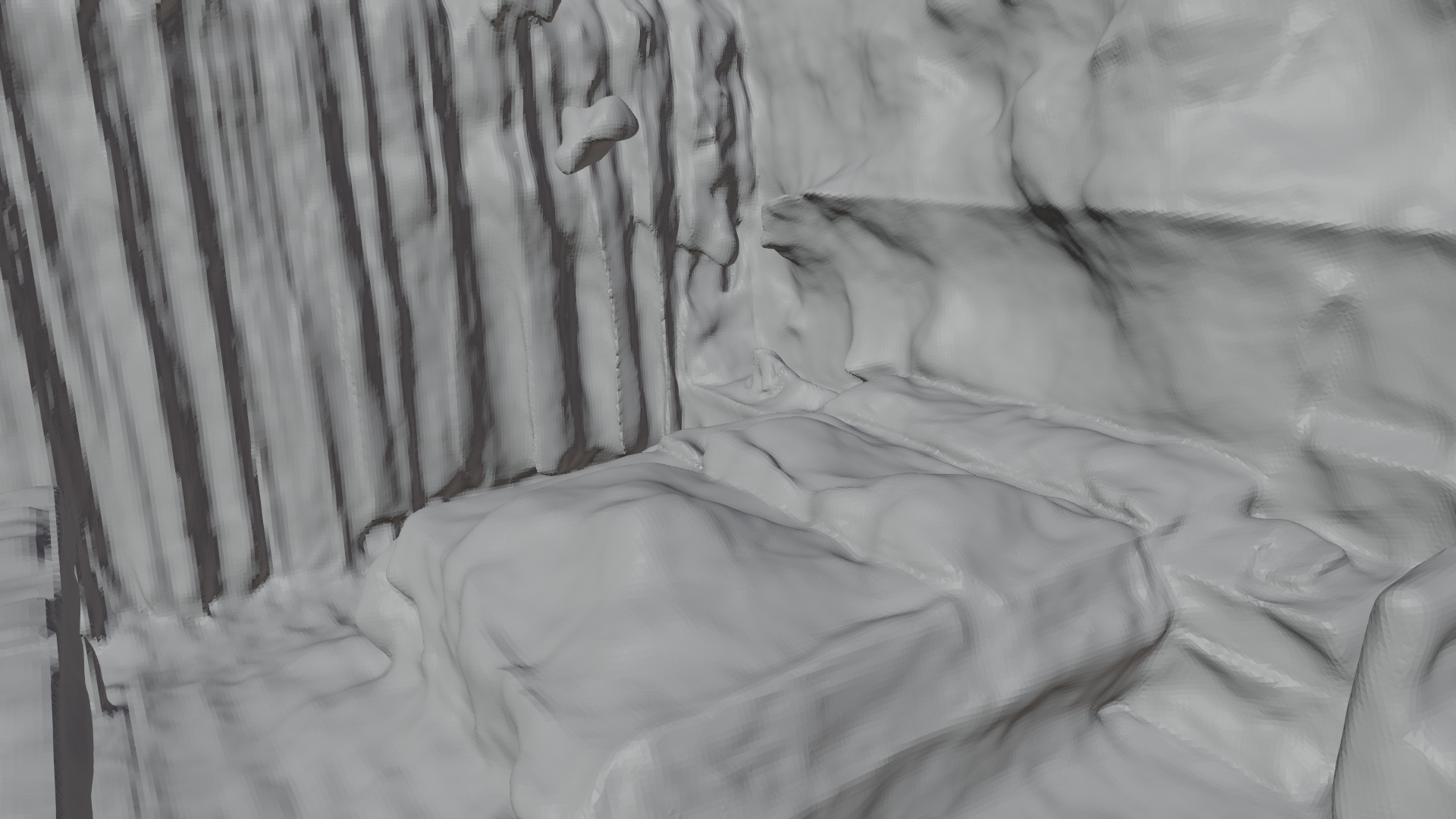}
\subcaption{\label{fig2}{\small NeuS \cite{neus}}}
\end{minipage}%
\begin{minipage}[t]{0.33\linewidth}
\centering
\includegraphics[width=0.95\linewidth]{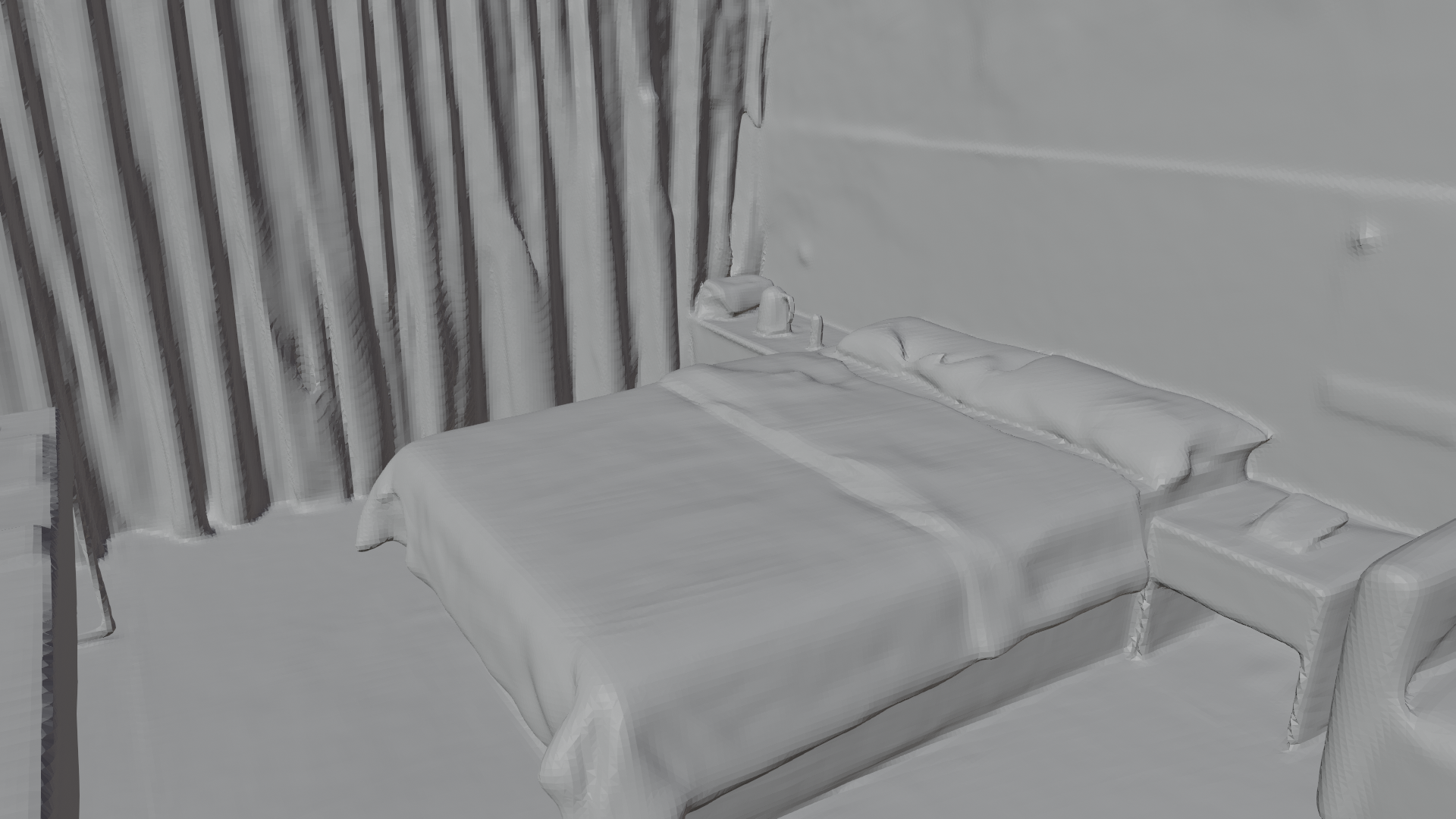}
\subcaption{\label{fig3}{\small Depth-NeuS (Ours)}}
\end{minipage}
\caption{\textbf{Comparison of the performance of NeuS and Depth-NeuS for indoor 3D reconstruction.} In the indoor scene of the example, our method's reconstruction effect is far superior to the traditional NeuS algorithm.}
\label{example}
\end{figure*}

Still, although we achieve the SOTA of the implicit surface representation, previous researches are struggling to discover efficient constraints for their training process. And another way to deal with this hindrance is to accelerate it by upgrading the hardware externally. We propose our Depth-NeuS, which doesn’t only work on refining the existing fundamentals, but also provide a scalable reconstruction solution for a variety of scenes. Depth-NeuS is able to reconstruct either narrow scene or large world scene with low-texture object surfaces. Also, our method can earn the highest PSNR score comparing with other models applying the SDF concepts or neural implicit representation. Initially, we demonstrate how the latest work, which solely relies on color field, has noticeable gap with ground-truth results. To maintain the outperformed results, we formulate a novel constraint on the multi-view synthesis task, which can refine the learning process by depth information. Moreover, our work integrate SDF projections for inverse image training, by which we develop a optimized depth-based geometry controlling for the reconstruction quality.Our method achieved good results in the reconstruction of indoor scenes as shown in Figure~\ref{example}, with significant improvements compared to NeuS.

In summary, our major contributions are:
We construct a new approach, Depth-NeuS, which can achieve scene-invariant reconstructions building on average computational capabilities and balanced memory consumption.
A tailored geometry constraint, interconnects the SDF representation for implicit surface and leverages the depth information.
An outperformed training methodology for neural implicit learning, which can both attain high score in accuracy, and minimize the complexity of prior knowledges of the scenes.

\section{Related works}

\noindent
\textbf{Implicit Neural Representation}
Implicit neural representation generally establishes new approaches to complete the scene representation and the understanding scene tasks. To better represent geometry in scenes, three works~\cite{IM-Net,Occupancy,deepsdf} concurrently propose the best practice of constructing grid-, point-, mesh-based representation for parameterizing. SRN~\cite{srn} established a differentiable renderer to train a neural network to represent a 3D coordinate in its feature. Followed the SRN’s groundwork for 3D reconstruction, a direct regression method~\cite{nerf} is constructed for RGB color and bring into the differentiable ray-matching renderer. Since then, the scene representation reaches the status-of-the-art, a bunch of follow-up works show the evolution in applying the NeRF’s solution.
After the accomplishment of scene representation, there’re several work that advances in allowing neural networks to understand the scene. The implicit representations of both geometry and appearance can be categorized by three supervision methods: 2D supervision, 3D supervision and hybrid supervision. Previous popular works provide solutions for learning features from inverse graphics, i.e. only using 2D supervision~\cite{DVR,pixelnerf}. Also, recent study has gained insights of novel views of the scene synthesis tasks. Specifically, to outperform the SOTA of novel view synthesis tasks, instant-NGP~\cite{instant} makes tradeoffs between memory consumption and computational capability, unfortunately it only achieves the latter. Our method, however, can speed up the training process while minimizing the complexity of supervision forms, thus optimizes the memory usage. \\
\noindent
\textbf{Geometric Constraints}
Traditional color-based 3D reconstruction performs well for small scenes and texture-rich areas. However, for larger scenes with sparse textures, the reconstruction process lacks explicit geometric constraints. Traditional multi-view geometry constraint methods~\cite{RC-MVSNet,multiview,geo-neus} can better capture the geometric information of 3D scenes through surface continuity and intersection geometry constraints. However, relying solely on multi-view geometry constraints still results in poor performance in texture-poor regions, such as doors and floors. To better handle point cloud rotation and evaluation, geometric constraints based on optical flow fields and camera poses~\cite{StructNeRF,GeCoNeRF} can better capture dynamic scenes and moving objects. These methods use vector projection of pixel points to obtain more accurate scene reconstruction results. Due to the demand for dense reconstruction in large scenes, Chen et al. proposed a method based on smooth neighboring frames and deep learning geometric constraints~\cite{dynamic} achieved good results in dense reconstruction without requiring multi-view inputs. Overall, geometry constraints based on RGB information are still not a true representation of the geometric clues in the image. To obtain detailed reconstruction results, significant challenges remain in the field of geometry constraints.\\
\noindent
\textbf{3D Reconstruction with Depth Information}
Depth information is crucial for 3D reconstruction as it provides the distance between every pixel and camera directly, which can be expressed as a depth value.  Making better use of depth information is a consistent work in 3D reconstruction. Truncated Signed Distance Function (TSDF)~\cite{TSDF,kinect1}, transform the depth information into a 3D voxel grid. TSDF is used to express the distance between every pixel and the surface of the object, which is stored in the pixel. It can deal with occlusion situations well, but it suffers from high complexity and memory consumption. Other methods~\cite{MVS,kinect2} predict depth maps for every image and use 3D information to complete the reconstruction. However, these multi-view stereo (MVS) methods have poor performance in low-texture areas. Some improvements make use of neural networks~\cite{geo-neus,neus} to optimize the expression of depth information, but all these depth information obtained through prediction have error in expressing the scene. Instead of predicting the depth map, We propose using real depth maps to support rendering in order to optimize detailed reconstruction.

\section{Method}
\begin{figure*}[t]
	\begin{center}
		\includegraphics[width=\linewidth]{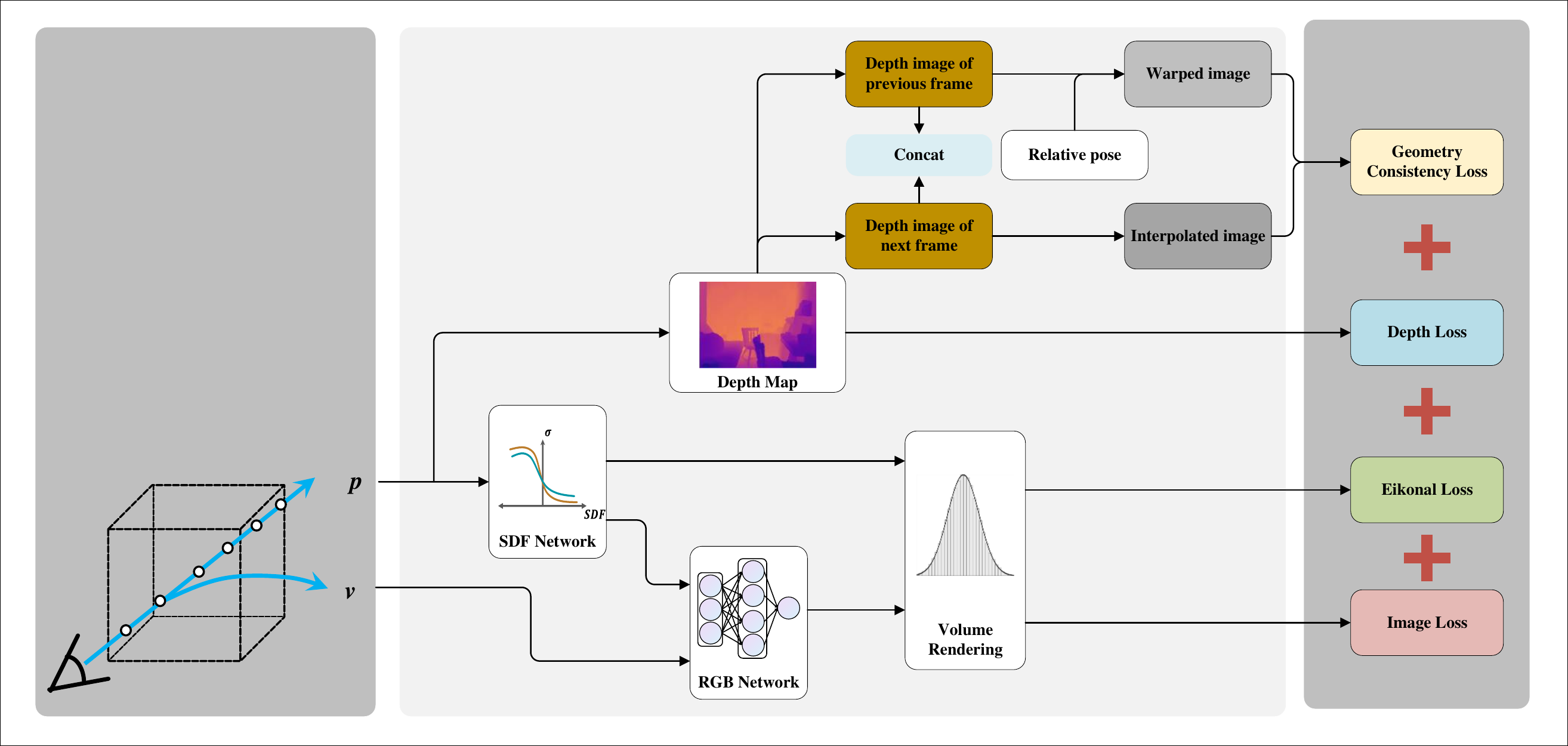}
	\end{center}
	\vspace{-8pt}
	\caption{\textbf{Overview of Depth-Neus.} Our work is based on neural implicit surface learning~\cite{neus}, and we additionally incorporate depth loss in Section~\ref{sub:depth loss} and geometry consistency loss in Section~\ref{sub:geomotry cons}, which significantly improves the reconstruction quality.}
	\label{fig:overview}
\end{figure*}
Given a set of multi-view RGB-D images of an object, our method aims to accurately reconstruct a high-fidelity surface representation encoded by implicit neural networks. In this article, we propose a new method called Depth-NeuS, as shown in Figure~\ref{fig:overview}. Building upon NeuS~\cite{neus}, we use signed distance and color fields to represent appearance and geometric features (Section~\ref{sub:color loss}). We optimize color changes in areas where depth information is unclear (Section~\ref{sub:depth loss}), and use geometric constraints to reconstruct precise geometry from real-world scenes (Section~\ref{sub:geomotry cons}).

\subsection{Learning Scene Representation Based on Volume Rendering}
\label{sub:color loss}
We adapt the neural surface representation under the theoretical guide from SDF fundamentals~\cite{siren}. This adaption will support our further analysis and experiments in latter sections. Each scene will be represented by a combination of MLPs: the geometry network and the color network. In the geometry network, given a specific 3D point x, it maps the point to a value of SDF function:
$$
f: \mathbb{R}^3 \rightarrow \mathbb{R}(\mathbf{x}, z)=\operatorname{SDF}(\mathbf{x})
$$
In the first MLP, we majorly trains the input coordinate x, to produce the most accurate distance, i.e. its value of signed distance function, accordingly. By the properties of SDF, our model can learn the surface as its zero-valued level set, that is:
$$
\mathcal{S}=\{\mathbf{x} \mid z(\mathbf{x})=0\}
$$
To efficiently construct the expected pixel-wise result from the original images, we follow the same rendering scheme proposed by NeRF~\cite{nerf}, we can covert the SDF values associated with the set of point x into density expression for performing volume rendering:
$$
\left\{G \mid \mathbf{x}_{\boldsymbol{i}}=\mathbf{0}+t_{\mathbf{r}}^i \mathbf{V}\right\}
$$
(i = 1,2,3…N, which is our sampled size along the ray direction v.)
Originally, the approximation of the color along with the ray origins at o, following the direction of v is:
$$
C(\mathbf{o}, \mathbf{v}, \mathbf{t})=\int_0^{+\infty} \omega(t) \rho(\mathbf{x}, \mathbf{z}, \mathbf{v}) d t
$$
While $\omega\left(t\right)$ is the learnable weight for volume rendering, $\rho\left(\mathbf{x},\mathbf{z},\mathbf{v}\right)$ is the expected color at point x. Discretizing the color approximation by sampling points, we can deduce it into:
$$
\hat{C}=\sum_{i=1}^N w\left(x_i\right) \hat{c}\left(x_i\right)
$$
\subsection{Depth Information Optimization}
\label{sub:depth loss}
Depth information offers significant advantages in reconstructing areas with rich textures, as it can accurately capture scene features. However, previous works that used depth information for reconstruction suffered from varying degrees of bias. Methods such as Neus' neural surface implicit representation obtain SDF information by learning RGB image features through a network\cite{neus}.
$$
\widehat{s d f}(\boldsymbol{p})=M L P(\boldsymbol{p})
$$
Where $\widehat{s d f}(\boldsymbol{p})$ represents the distance between spatial point $\boldsymbol{p}$ and the object surface, and $MLP$ is the network. This method of predicting by learning data features inevitably has errors. 

In Geo-NeuS~\cite{geo-neus}, interpolation and SFM~\cite{sfm} supervision are used to assist the prediction. However, the effect of solving the error problem has not achieved the expected results due to the inherent defects of these auxiliary methods. For the MVS~\cite{2016mvs} method, the 3D information of the scene is attempted to be restored through images by matching the same pixel points from different viewpoints: 
$$
\boldsymbol{x_1}^T K \boldsymbol{x_2}=0
$$
Where $\boldsymbol{x_1}$ and $\boldsymbol{x_2}$ represent pixel points under different perspectives, and $K$ represents the fundamental matrix. Due to occlusions and camera noise, this method is inevitably prone to errors, and visualized depth maps generated by this method may show distorted shapes and positions of certain objects. To improve upon this method, MVSNet~\cite{mvsnet} attempts to optimize the reference image and depth map as inputs, resulting in significant improvements in effectiveness.
Similarly, we aim to improve the accuracy of predicted images by utilizing real depth images as input based on the neural surface implicit representation method. For each pixel, the network can predict its depth and compare it with the depth in the real depth image to supervise the expression of scene depth information. By optimizing the real depth image, we can further enhance surface quality and increase the contribution of depth information to 3D reconstruction.

\subsection{Geometry Constraints Based on Scale-Invariant Depth}
\label{sub:geomotry cons}
For large-scale 3D reconstruction tasks, constraining geometric features through color information often yields poor results. This is because RGB information is missing in large scenes and there is a lack of texture in areas such as walls and floors. We propose a geometry constraint method based on scale-consistent depth, which projects the previous image onto the current image using the input RGB-D image and camera pose to obtain a consistent scale prediction. Compared with multi-view geometry constraint methods~\cite{ACMM}, our method uses more direct geometric clues. Our geometry constraint consists of two parts: photometric consistency constraint and geometry consistency constraint. The photometric loss is:
$$
\mathcal{L}_{\text {photo }}=\frac{1}{|M|} \sum_{p \in M}\left|I_2(p)-I_2^{\prime}(p)\right|
$$
Where $I_2^{\prime}$ represents the RGB image distorted by projecting $I_1$, $M$ represents the valid points successfully projected from $I_1$ to $I_2$, and $|M|$ represents the number of valid points.
And the geometric loss is:
$$
\mathcal{L}_{\text {geo }}=\frac{1}{|N|} \sum_{p \in N} \frac{\left|D_2^1(p)-D_2^{\prime}(p)\right|}{D_2^1(p)+D_2^{\prime}(p)}
$$
Where $D_2^1$ represents the depth image obtained by distorting the projection of $D_1$, $N$ represents the effective points successfully projected from $D_1$ to $D_2$, while $|N|$ represents the number of effective points.

\subsection{Loss Functions}
During rendering colors from a specific view, our total loss is:
$$
\mathcal{L}=\mathcal{L}_{\text {img }}+\alpha \mathcal{L}_{\text {eikon }}+\beta \mathcal{L}_{\text {depth }}+\gamma \mathcal{L}_{G C}
$$

Where $\mathcal{L}_{\text {img }}$ is the color field loss, defined as:
$$
\mathcal{L}_{\text {img }}=\sum_{\mathbf{r} \in G}\|\hat{C}(\mathbf{r})-C(\mathbf{r})\|_1
$$

Where $C(\mathbf{r})$ is the real-world color along the ray $\mathbf{r}$, $G$ is the series of rays go into the sampled points. We choose $L_1$ loss because it has better robustness for large scenes.

$\mathcal{L}_{\text {eikon }}$ is Eikonal loss proposed by previous research to assist the regularization of SDF values on sampled points:
$$
\mathcal{L}_{\text {eikon }}=\sum_{\mathbf{x} \in U}\left(\left\|\nabla_{\mathbf{x}} z(\mathbf{x})\right\|_2-1\right)^2
$$

Where $U$ stands for the set of sampled points $\mathbf{x}$ derived from surface and random areas.

$\mathcal{L}_{\text {depth }}$ is Depth loss, defined as:
$$
\sum_{p \in U}|D(p)-\hat{D}(p)| * \text { mask }
$$

Where $D(p)$ is the real depth of the sampled point. $U$ stands for the set of sampled points. $\text { mask }$ is a boolean value which can be 0 or 1, which used to limit the error value to a portion of the target point cloud, rather than the entirety, for more accurate error calculation.

$\mathcal{L}_{G C}$ is geometry consistency loss, defined as:
$$
\mathcal{L}_{G C}=\mathcal{L}_{\text {photo }}+\mathcal{L}_{\text {geo }}
$$

Where $\mathcal{L}_{\text {photo }}$ stands for photometric loss and $\mathcal{L}_{\text {geo }}$ stands for geometric loss.

Through multiple experiments, we determined that the optimal parameters for the loss function are $\alpha=0.7, \beta=1.0, \gamma=0.5$.

\section{Experiment}
\subsection{Experiment Settings}

\noindent
\textbf{Datasets.}
To test the performance of our method and fully leverage the advantages of depth information in texture-rich areas~\cite{depth}, we chose the ScanNet dataset. ScanNnet~\cite{scannet} contains RGB-D images and corresponding 3D geometry and semantic labels for more than 1,500 indoor scenes with rich texture information. At the same time, to test the performance in low-texture areas such as walls and verify the effectiveness of geometric constraints, instead of only choosing the rich texture areas of the room, we performed whole-room scale reconstruction. We selected six scenes that have both rich texture and low-texture areas. For each scene, we selected 200-500 images. The number of photos selected for each scene is related to the length of the video, and the longer the video, the more photos are selected.

\noindent
\textbf{Network architecture.}
Our network architecture is similar to NeuS~\cite{neus}, which models the SDF field and color function using an MLP model with eight hidden layers of size 256. To improve the model's performance and generalization, we applied spherical initialization and positional encoding, with reference to DeepSDF~\cite{deepsdf} and Ha-NeRF~\cite{hanerf}. For volume rendering, we used a hierarchical sampling strategy, iterating over 512 rays and sampling points for each ray following NeuS's~\cite{neus} approach. During training, we used the RMSProp~\cite{rmsprop} optimizer for its good performance on sparse data. We initialized the learning rate to 0.01 and decreased it as necessary in subsequent iterations. We trained for 160k iterations on a single NVIDIA RTX3090 GPU, which took about four hours. However, our method is significantly faster than other 3D reconstruction methods. 

\noindent
\textbf{Baseline.}
We compared our method with the following types of methods: (1) TSDF-based depth reconstruction method: Atlas\cite{atlas}; (2) Traditional MVS method: COLMAP\cite{2016mvs}; and (3) Neural surface implicit expression method: NeRF\cite{nerf}, Neus\cite{neus}, VolSDF\cite{volsdf}, and Geo-Neus\cite{geo-neus}. For the COLMAP method,we used the Delaunay triangulation method\cite{delaunay} and also estimated the normal vectors while restricting the maximum edge length of the triangle.These steps help us export a more refined mesh.

\noindent
\textbf{Evaluation metrics.}
To quantitatively compare different 3D reconstruction methods, we evaluated the reconstructed mesh using five metrics defined in Atlas~\cite{atlas}: accuracy, completeness, precision, recall, and F-score. F-score is typically considered the overall metric.

\subsection{Comparisons}
\begin{figure*}[htbp]
    \captionsetup[subfigure]{labelformat=empty}
    \centering
    \hspace*{\fill}
     \rotatebox{90}{\scriptsize{~~~~~~~~Scene 580}}
    \hspace*{\fill}
     \begin{subfigure}[b]{0.24\linewidth}
         \centering
         \includegraphics[width=\textwidth]{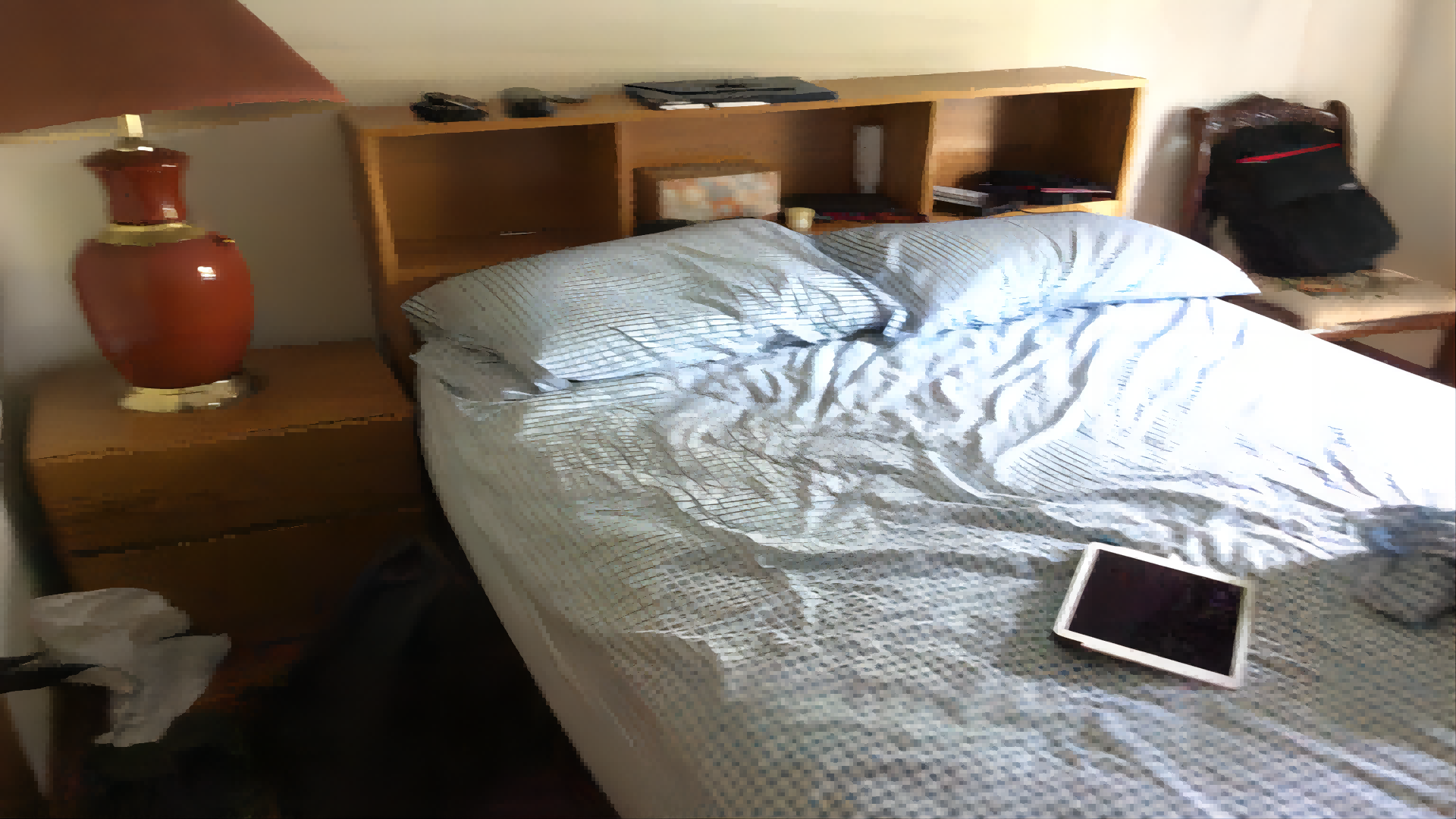}
     \end{subfigure}
     \begin{subfigure}[b]{0.24\linewidth}
         \centering
         \includegraphics[width=\textwidth]{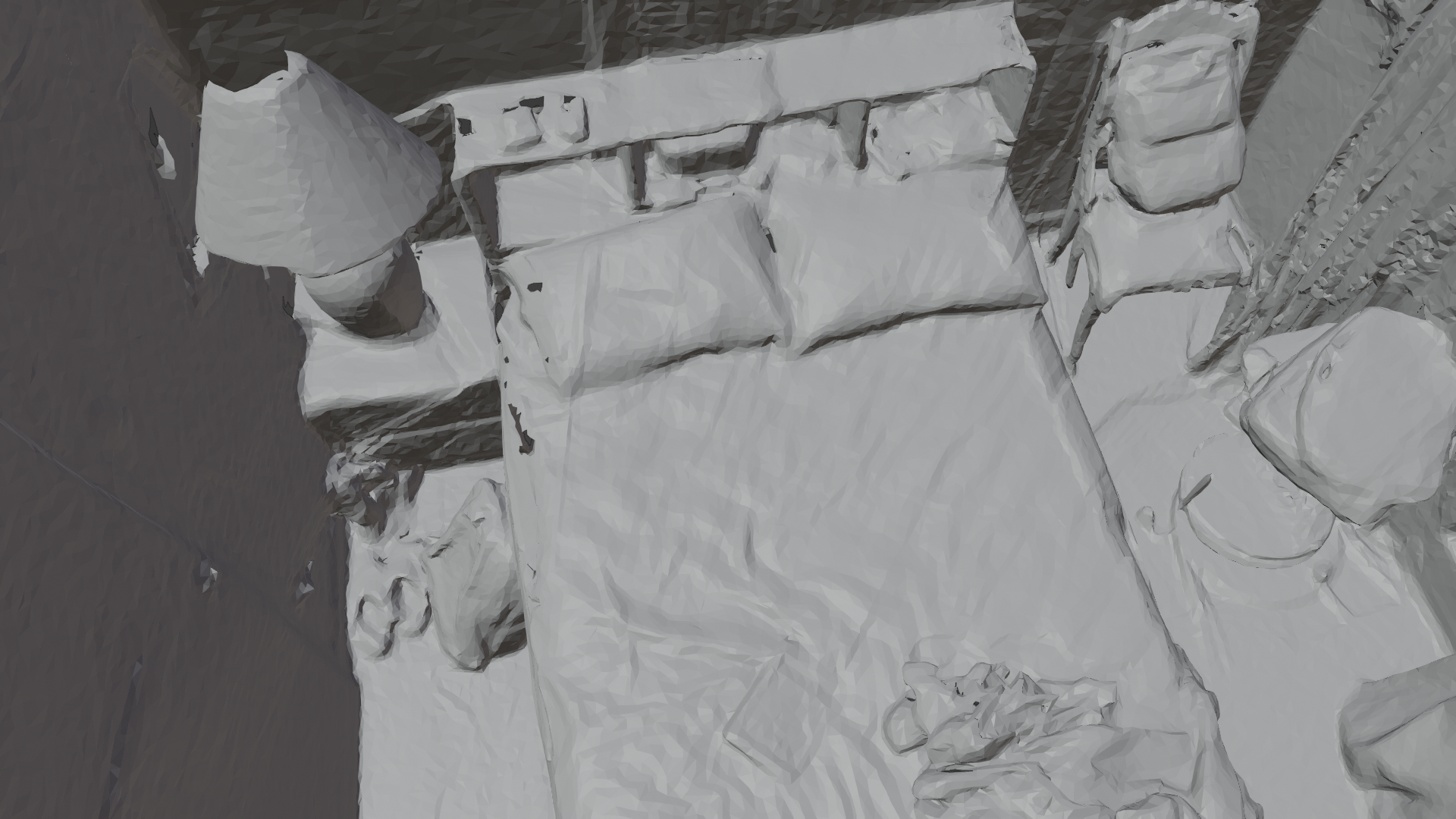}
     \end{subfigure}
     \begin{subfigure}[b]{0.24\linewidth}
         \centering
         \includegraphics[width=\textwidth]{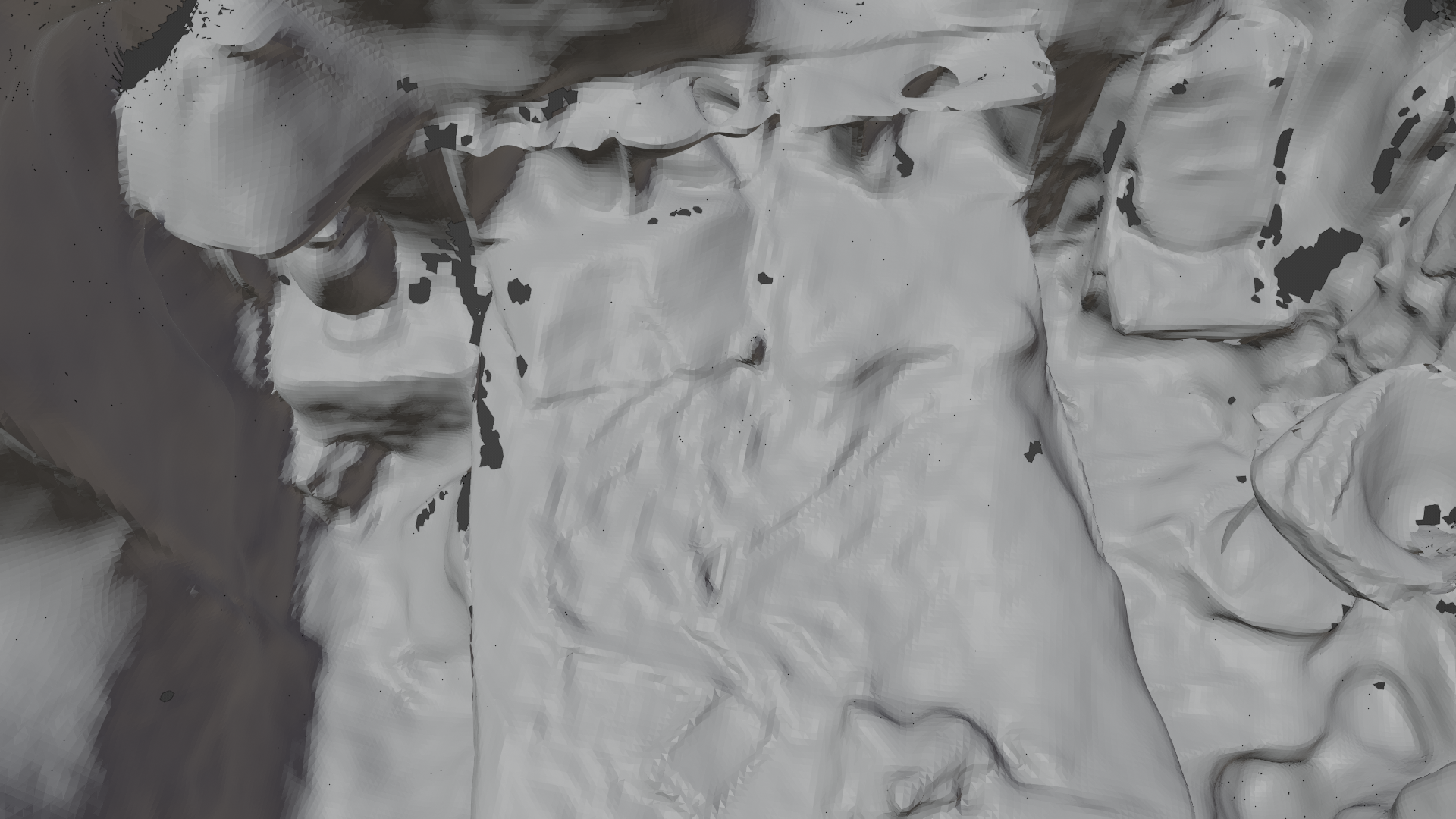}
     \end{subfigure}
     \begin{subfigure}[b]{0.24\linewidth}
         \centering
         \includegraphics[width=\textwidth]{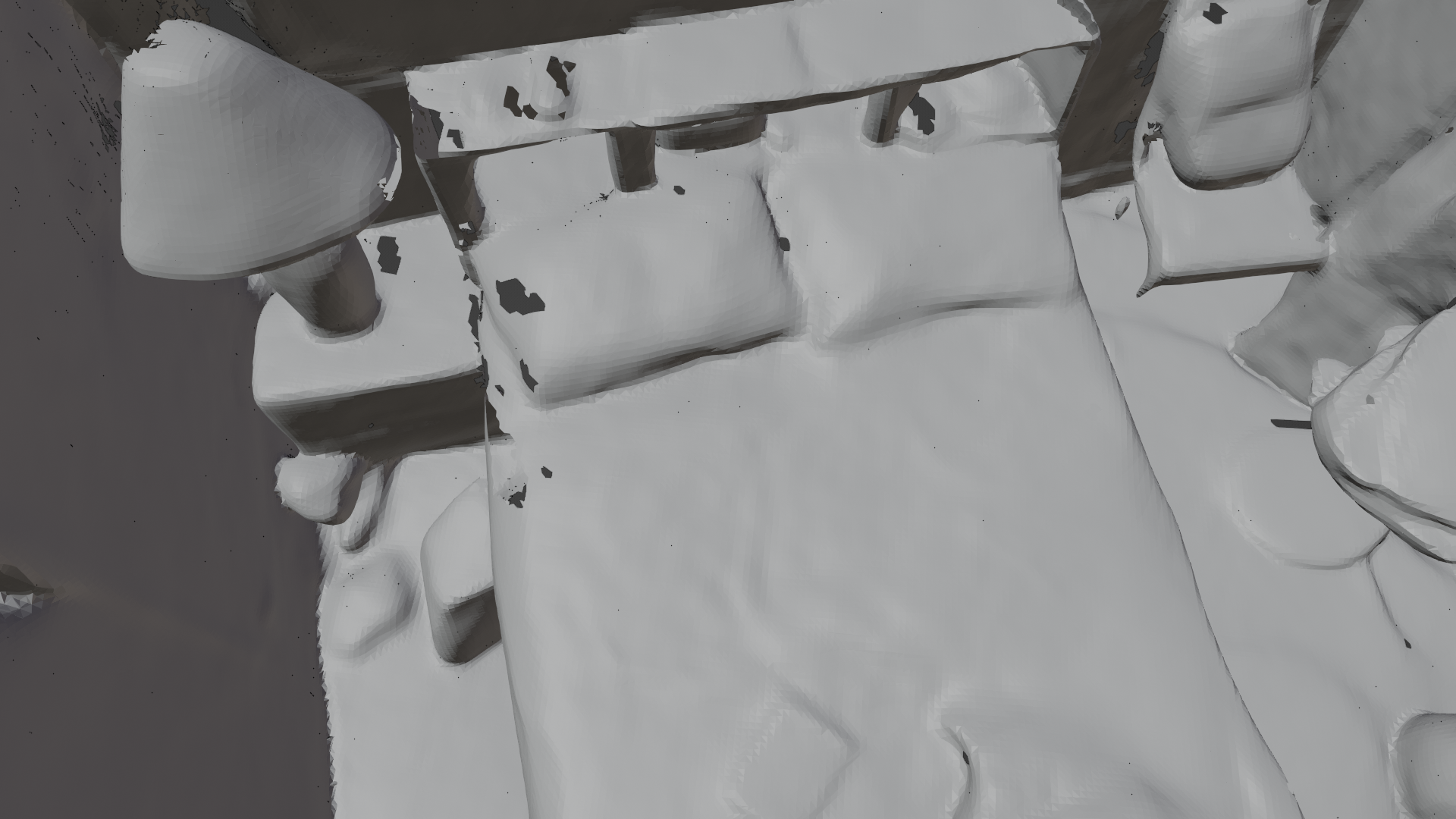}
     \end{subfigure}
    \hspace*{\fill}
    
    \vspace{0.5mm}
    
     \hspace*{\fill}
     \rotatebox{90}{\scriptsize{~~~~~~~Scene 603}}
    \hspace*{\fill}
     \begin{subfigure}[b]{0.24\linewidth}
         \centering
         \includegraphics[width=\textwidth]{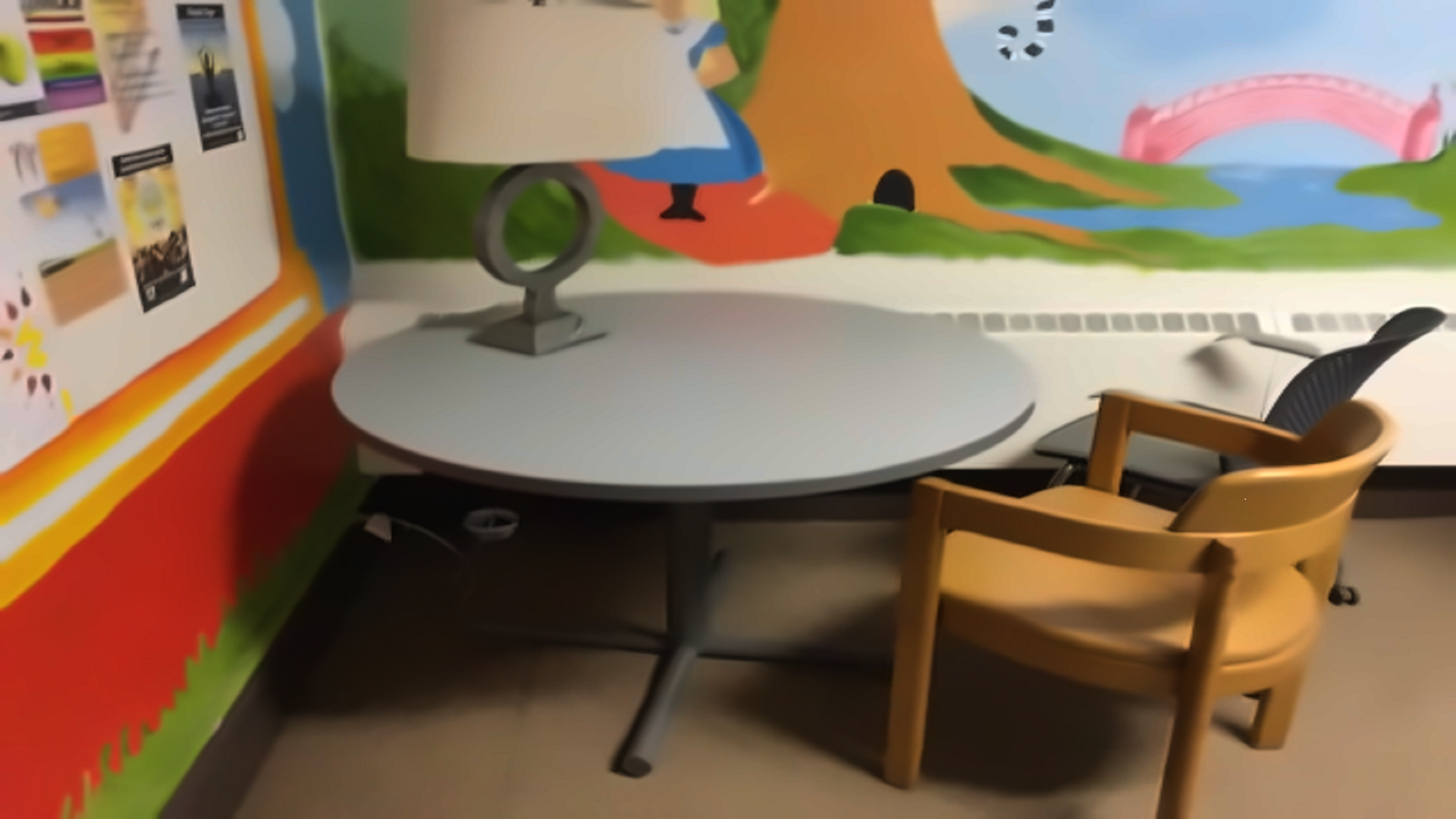}
     \end{subfigure}
     \begin{subfigure}[b]{0.24\linewidth}
         \centering
         \includegraphics[width=\textwidth]{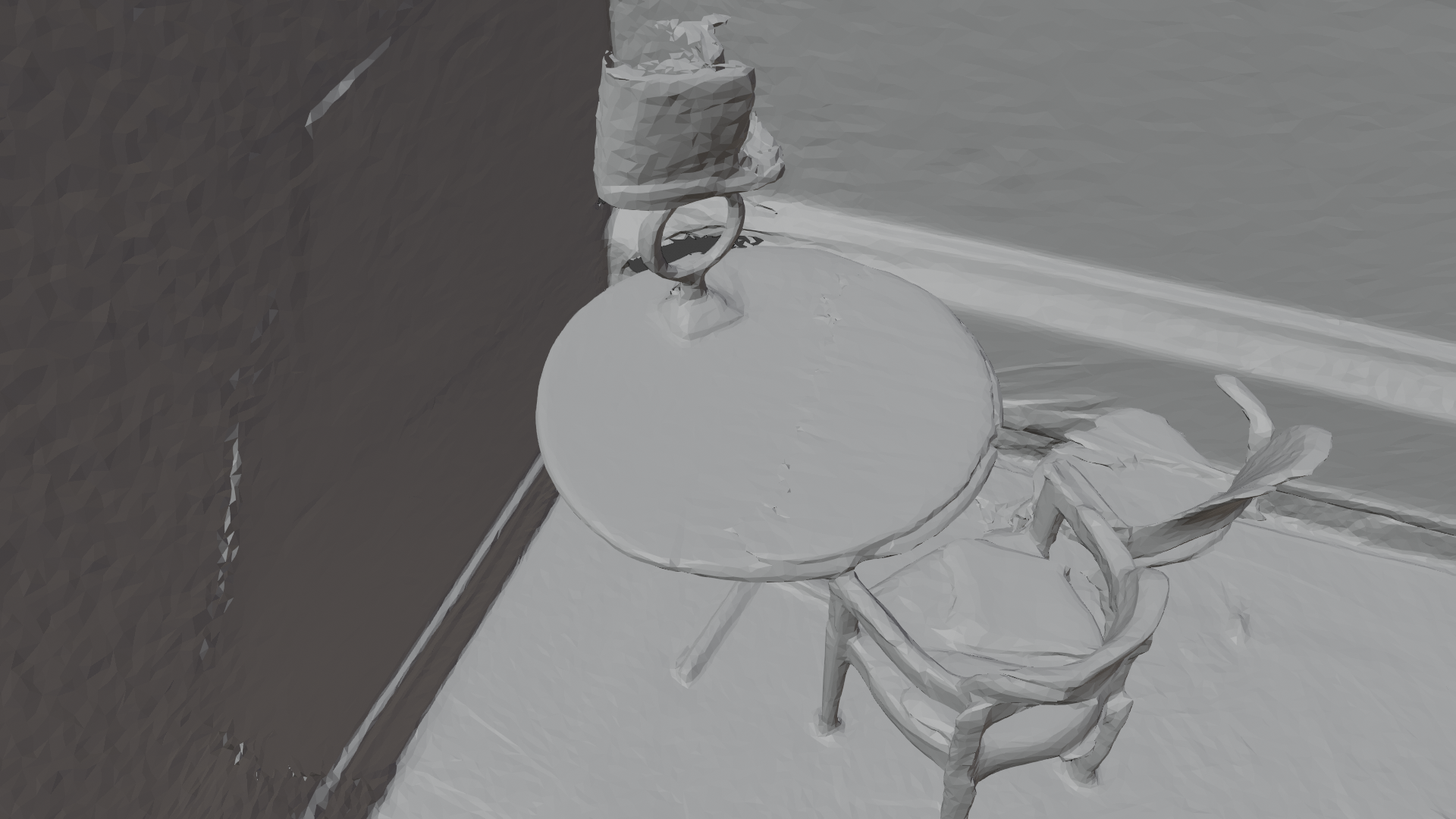}
     \end{subfigure}
     \begin{subfigure}[b]{0.24\linewidth}
         \centering
         \includegraphics[width=\textwidth]{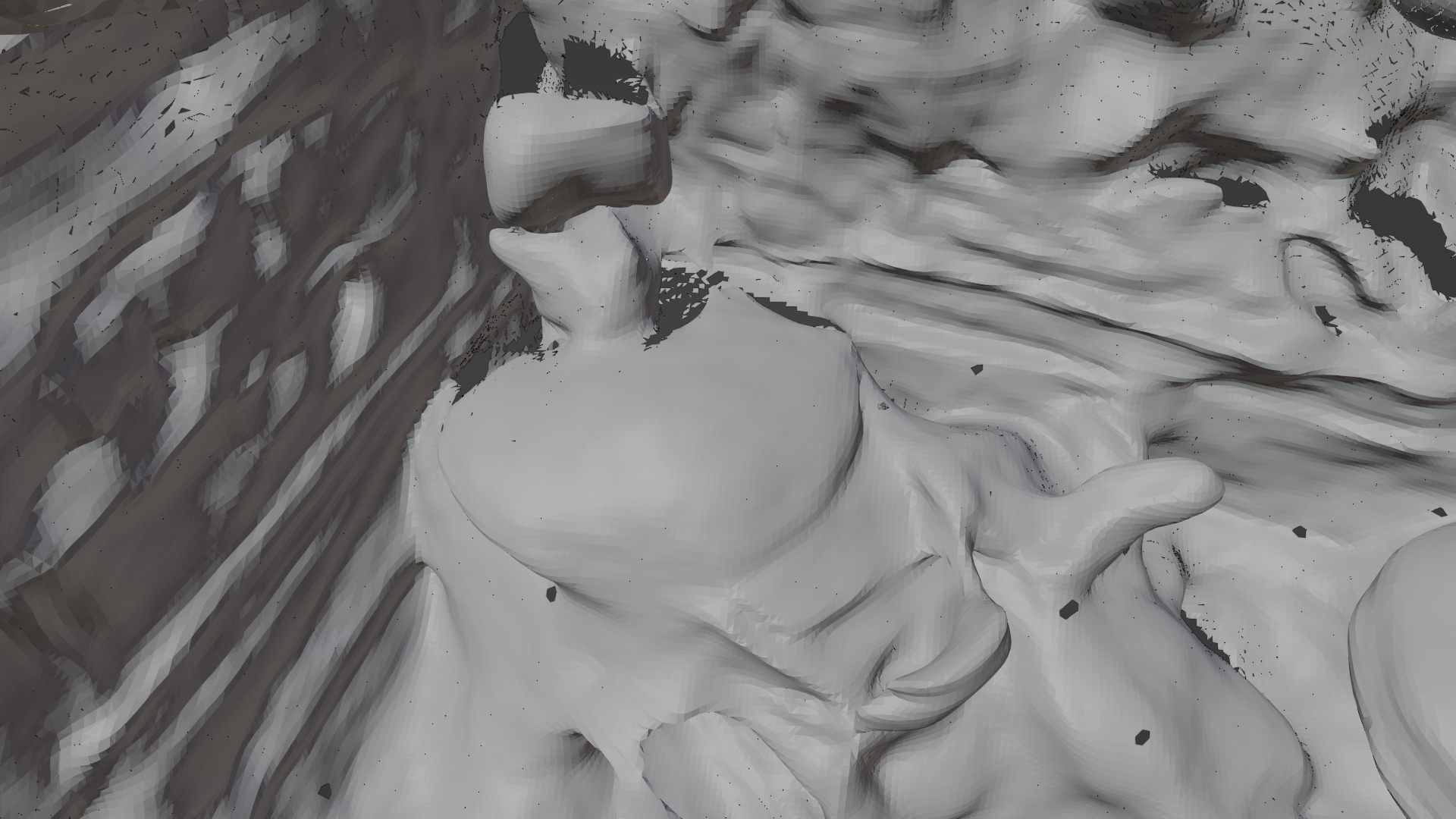}
     \end{subfigure}
     \begin{subfigure}[b]{0.24\linewidth}
         \centering
         \includegraphics[width=\textwidth]{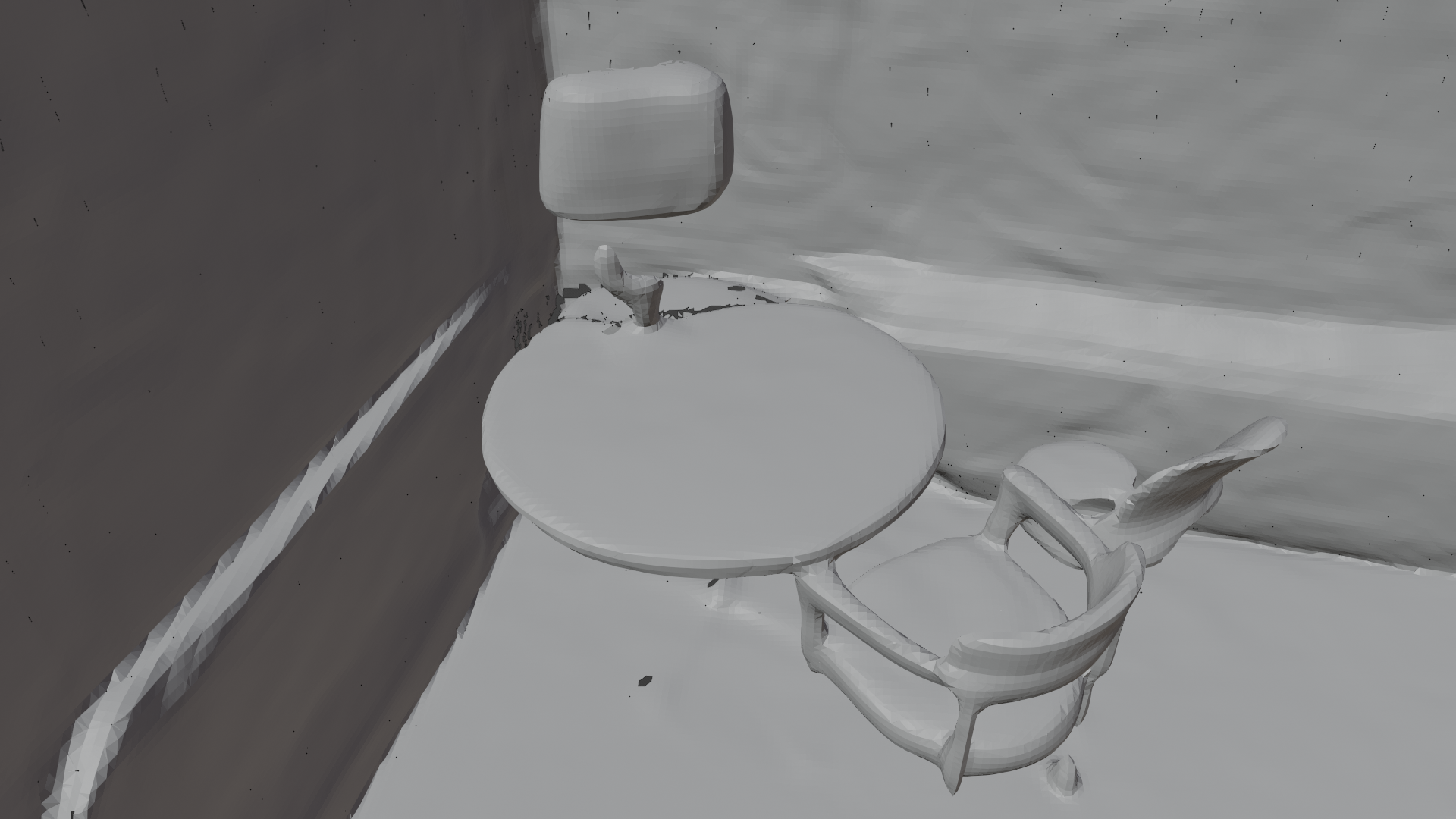}
     \end{subfigure}
    \hspace*{\fill}
    
    \vspace{0.5mm}
    
     \hspace*{\fill}
     \rotatebox{90}{\scriptsize{~~~~~~~~~~~~~~~~~Scene 625}}
    \hspace*{\fill}
     \begin{subfigure}[b]{0.24\linewidth}
         \centering
         \includegraphics[width=\textwidth]{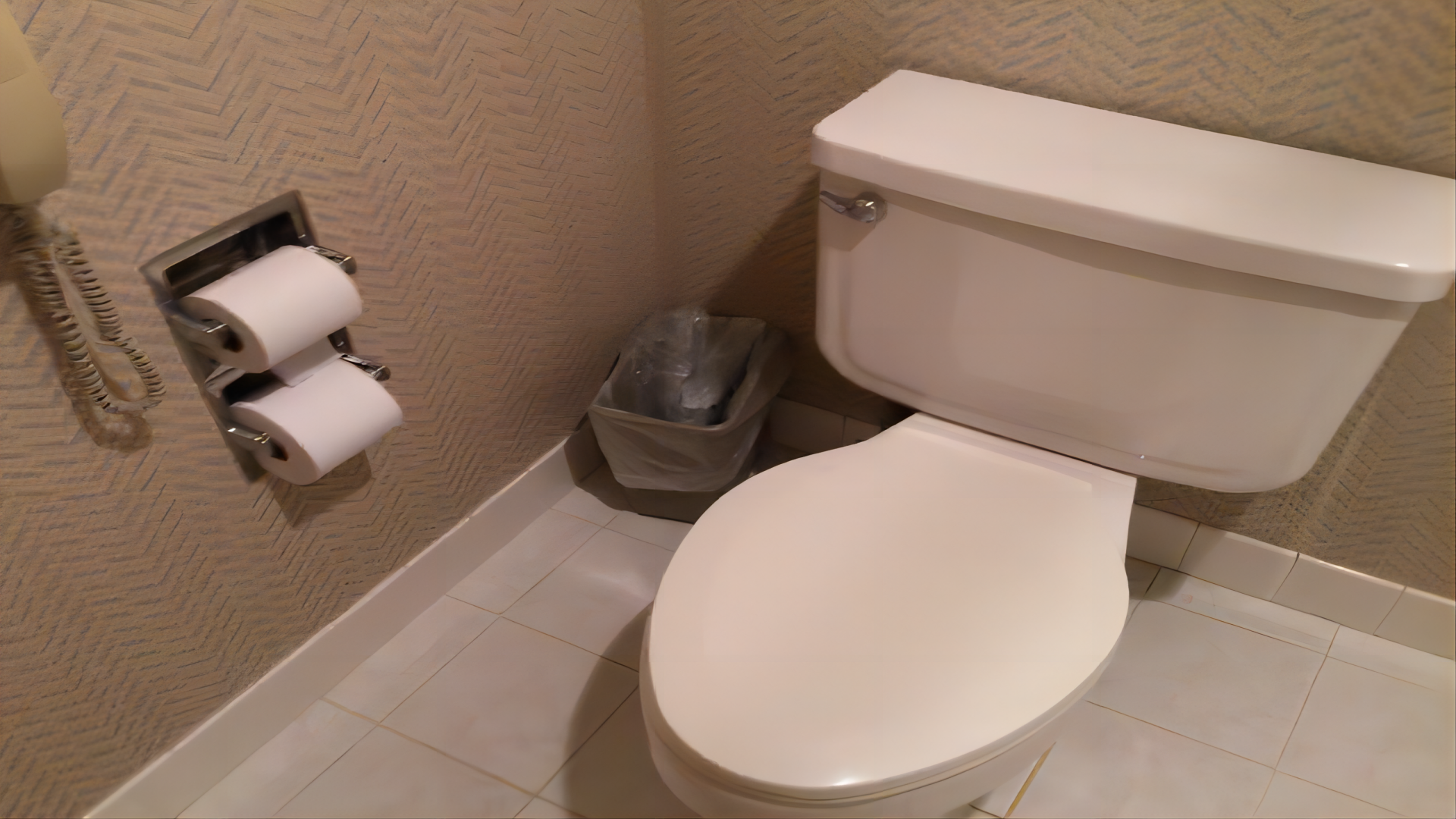}
        \caption{Reference Image}
     \end{subfigure}
     \begin{subfigure}[b]{0.24\linewidth}
         \centering
         \includegraphics[width=\textwidth]{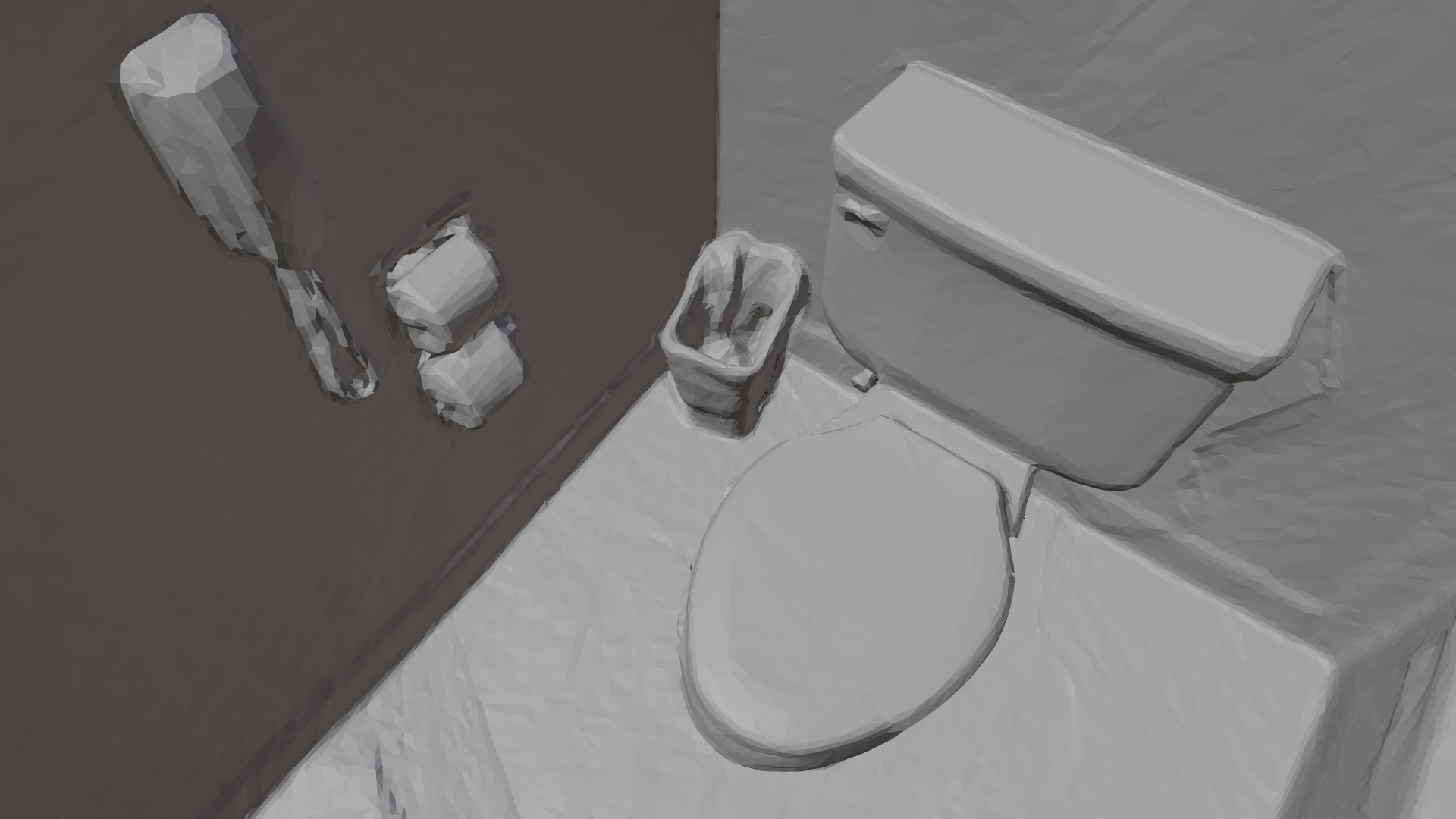}
        \caption{GT}
     \end{subfigure}
     \begin{subfigure}[b]{0.24\linewidth}
         \centering
         \includegraphics[width=\textwidth]{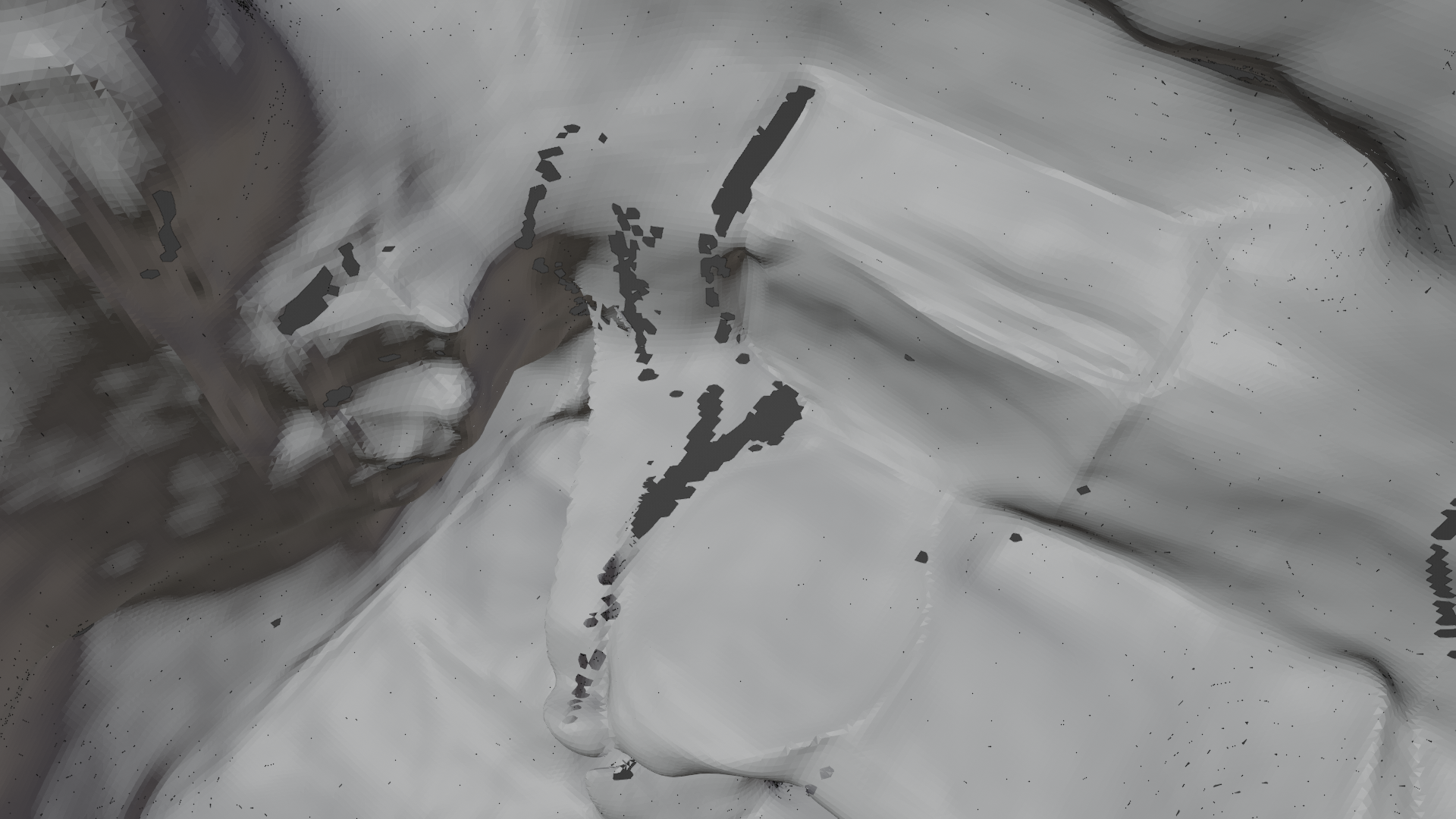}
        \caption{NeuS~\cite{neus}}
     \end{subfigure}
     \begin{subfigure}[b]{0.24\linewidth}
         \centering
         \includegraphics[width=\textwidth]{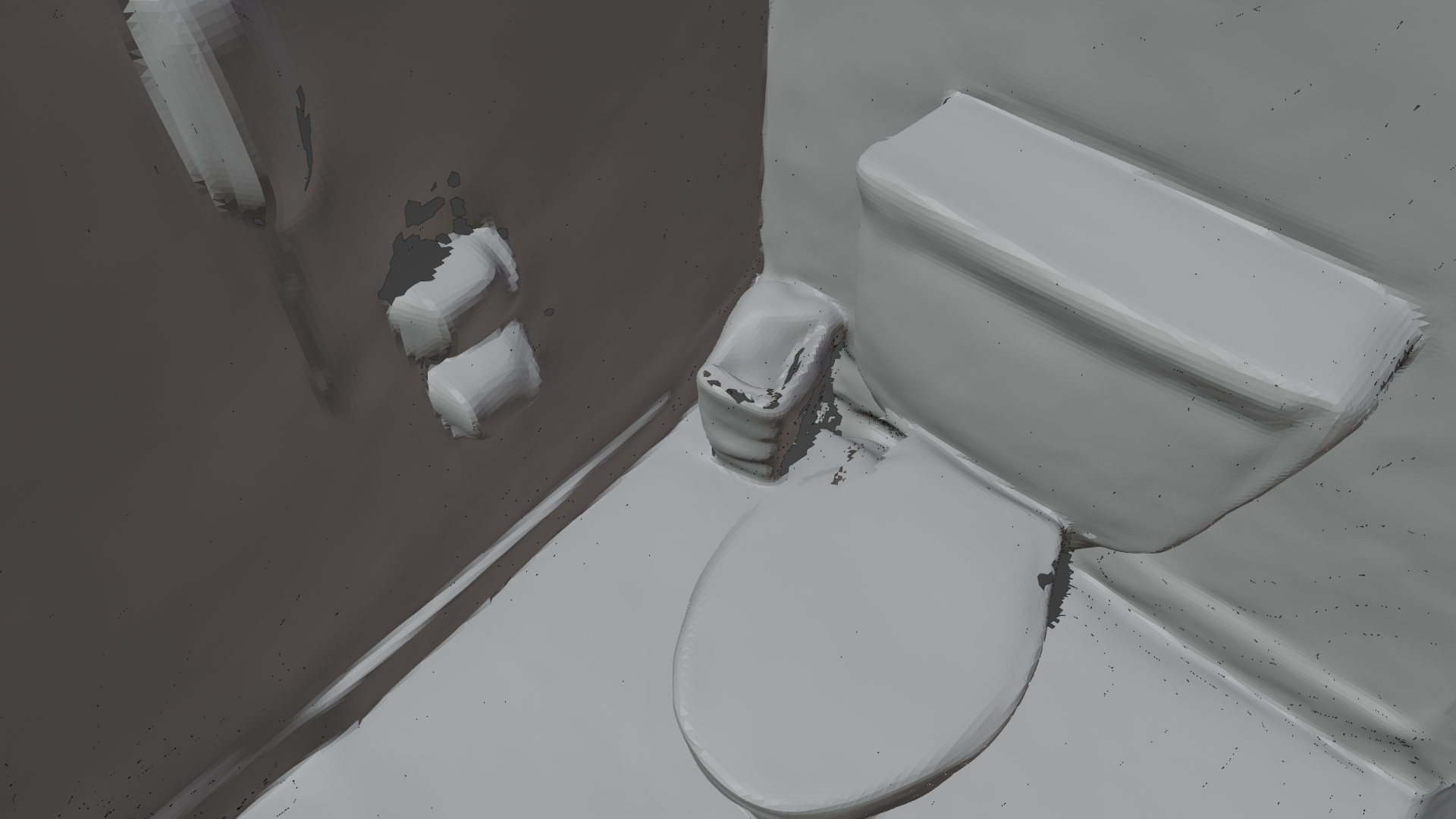}
        \caption{Depth-NeuS (Ours)}
     \end{subfigure}
    \hspace*{\fill}
    \caption{\textbf{Comparisons on surface reconstruction in ScanNet dataset.} Our reconstruction results are much better than other works based on Implicit Neural Representation, restoring most of the details in large indoor scenes and performing well in low-texture areas.}
    \label{fig:comparison_scannet}
\end{figure*}

\begin{table*}[htbp]
\begin{center}
\resizebox{.7\textwidth}{!}{
\begin{tabular}{c|cccccc|c}
 Scene &580&603&616&617&625&771&means\\
\hline 
NeRF~\cite{nerf} & 24.38 & 23.65 & 23.85 & 24.73 & 30.88 & 29.50 & 26.00\\
VolSDF~\cite{volsdf} & 26.62 & 25.53 & 24.92 & 26.96 & 32.01 & 29.43 & 27.58\\
NeuS~\cite{neus} & 26.59 & 26.46 & 24.80 & 27.19 & 31.86 & 30.02 & 27.82\\
Geo-NeuS~\cite{geo-neus} & \underline{27.90} & \textbf{26.74} & \textbf{25.03} & \underline{27.89} & \underline{32.25} & \underline{30.75} & \underline{28.43}\\
Depth-NeuS (Ours) & \textbf{27.96} & \underline{26.48} & \underline{25.92} & \textbf{28.47} & \textbf{34.71} & \textbf{30.84} & \textbf{29.06}\\
\end{tabular}}
\end{center}
\caption{\textbf{Qualitative results on ScanNet dataset in terms of PSNR.} Although the performance of our method on the two scenes is not optimal, according to the final means, our method is better than the other methods.}
\label{psnr}
\end{table*}
To assess the quality of surface geometry reconstruction, we evaluate the 3D geometry metrics of our method with state-of-the-art methods. Table~\ref{psnr} shows the PSNR metrics of different methods~\cite{geo-neus,nerf,neus,volsdf} on six scenes of the ScanNet dataset, representing rendering performance for different scenes. From the table, it can be seen that our method performs better in rendering, with an average PSNR score higher than other methods. The reason why our method can achieve a higher PSNR score is that we utilize depth information to provide geometric reference for accurate and complete reconstruction. In contrast, other methods that rely on color fields or volume rendering may be affected by geometric inaccuracies or incomplete details, resulting in lower PSNR scores. Qualitative results in Fig~\ref{fig:comparison_scannet} show that our work yields more consistent surface quality and more accurate detail than other works. Table~\ref{table01} shows the average value of each evaluation indicator for different works~\cite{geo-neus,nerf,atlas,2016mvs,neus,volsdf} in the six scenarios we selected. From Table~\ref{table01}, we can see that our method outperforms other neural implicit surface learning methods in all metrics except recall rate. Specifically, our method performance is over 50$\%$ better than the traditional NeuS~\cite{neus}. Our method performs well in the reconstruction of the ScanNet dataset. Compared to other methods, we use a more direct geometric reference retrieval, that is, depth information, which allows us to obtain accurate and complete reconstruction results in large or small scenes, areas with rich textures or areas with texture deficiencies. In addition, we also use some other techniques such as geometric constraints and smoothness constraints, which make our reconstruction results more realistic and convincing. Furthermore, our method is not only applicable to the ScanNet dataset, but also to other scene or object reconstruction tasks with depth information input, with high versatility and scalability.\\
Geo-NeuS~\cite{geo-neus} outperforms other methods based on volume rendering and Implicit Neural Representation, due to its use of multi-view geometry constraints. However, its geometric constraints are based on color fields, making it difficult to accurately restore geometric features in areas with little color variation. In contrast, our depth-based geometric consistency constraint has a greater constraint force on geometry, making it perform better in terms of F-score compared to other works.
\begin{table*}[htbp]
\begin{center}
\setlength{\tabcolsep}{1.5mm}{
\begin{tabular}{lccccc}
\toprule
& Acc$\downarrow$ & Comp$\downarrow$ & Prec$\uparrow$ & Recall$\uparrow$ & F-score$\uparrow$ \\
\midrule
COLMAP~\cite{2016mvs} & 0.078 & 0.081 & 0.563 & 0.587 & 0.548  \\
NeRF~\cite{nerf} & 0.231 & 0.103 & 0.321 & 0.294 & 0.256  \\
VolSDF~\cite{volsdf}  & 0.209 & 0.099 & 0.311 & 0.255 & 0.345  \\
NeuS~\cite{neus} & 0.143 & 0.137 & 0.474 & 0.495 & 0.473  \\
Atlas~\cite{atlas} & 0.203 & \underline{0.065} & 0.542 & 0.628 & 0.577  \\
Geo-NeuS~\cite{geo-neus} & \underline{0.076} & 0.081 & \underline{0.652} & \textbf{0.694} & \underline{0.663}  \\
Ours  & \textbf{0.049} & \textbf{0.051} & \textbf{0.743} & \underline{0.685} & \textbf{0.712}  \\
\bottomrule
\end{tabular}
}
\end{center}
\caption{\textbf{Quantitative comparisons of 6 scenes on ScanNet.} We selected 6 scenes from the ScanNet dataset and compared our method with other state-of-the-art methods.}
\vspace{-0.2cm}
\label{table01}
\end{table*}
\subsection{Ablation Studies}
To evaluate the impact of the three loss functions we used on the model performance, we conducted ablation experiments following the configurations below. We considered the following three cases: (1) \textbf{NeuS:} the traditional NeuS, which only uses color field loss and SDF gradient loss. (2) \textbf{NeuS-D:} adding depth loss on the basis of traditional NeuS. (3) \textbf{NeuS-D-G:} adding depth loss and geometric consistency loss on the basis of traditional NeuS.\\
According to Table~\ref{table:ablation}, for the ScanNet scene we have chosen, using only color information is not sufficient. Due to the large size of the scene and incomplete color information, this can lead to poor reconstruction results. With the addition of depth information, the accuracy and recall of the model are significantly improved. Adding geometric consistency loss ensures more complete surface features are restored in low-texture areas, leading to optimal model performance. In 3D reconstruction tasks for larger scenes, RGB information is often lost, thus using depth information to optimize reconstruction tasks is necessary.

\begin{table}[ht]
\centering
\setlength{\tabcolsep}{1.5mm}{
\begin{tabular}{lccccc}
\toprule
& Acc$\downarrow$ & Comp$\downarrow$ & Prec$\uparrow$ & Recall$\uparrow$ & F-score$\uparrow$ \\
\midrule
NeuS     & 0.143 & 0.137 & 0.474 & 0.495 & 0.473  \\
NeuS-D   & 0.089 & 0.092 & 0.646 & 0.601 & 0.623  \\
NeuS-D-G & \textbf{0.049} & \textbf{0.051} & \textbf{0.743} & \textbf{0.685} & \textbf{0.712}  \\
\bottomrule
\end{tabular}
}
\caption{\textbf{Ablation studies of 6 scenes on ScanNet.} According to our evaluation criteria, we found that the model achieves the best performance when using both deep loss and geometric consistency loss simultaneously.}
\label{table:ablation}
\vspace{-0.2cm}
\end{table}

\section{Conclusion}
This article introduces a new neural implicit surface learning method based on optimized depth information. The key idea is to use the input depth map to optimize and constrain geometric reconstruction. First, we provide theoretical analysis showing that 3D reconstruction based on colour fields has errors. Secondly, based on this theory, we propose a depth loss function and a scale-invariant depth based geometric consistency loss function. The experiments show that in low-texture and colour-invariant areas, high-quality reconstruction results can be produced in this way. Compared to traditional neural implicit surface learning methods, model performance has increased by more than 50$\%$. Although our method significantly improves reconstruction quality, its training convergence speed is still insufficient to meet industrial demands. Recent work~\cite{instant} uses hash coding to learn neural implicit surfaces for more significant effects. In the future, we will attempt to integrate acceleration methods into our network framework to achieve higher model performance.
\section*{Acknowledgement}
This work was supported in part by the National Natural
Science Foundation of China (Grant No. 61834006, 62025404), and in part by the Strategic Priority Research Program of Chinese Academy of Sciences (Grant No. XDB44030300, XDB44020300).

{\small
\bibliographystyle{ieee_fullname}
\bibliography{egbib}
}

\end{document}